\documentclass[10pt, conference]{IEEEtran}
\makeatletter
\def\ps@headings{%
\def\@oddhead{\mbox{}\scriptsize\rightmark \hfil \thepage}%
\def\@evenhead{\scriptsize\thepage \hfil \leftmark\mbox{}}%
\def\@oddfoot{}%
\def\@evenfoot{}}
\makeatother
\usepackage[utf8]{inputenc}

\usepackage[english]{babel}

\usepackage[noadjust]{cite}
\usepackage{amsmath}
\usepackage{tikz}
\usepackage{hyperref}
\usepackage{pgfplots}
\pgfplotsset{compat=1.15}
\usepackage{mathrsfs}
\usetikzlibrary{arrows}
\usepackage{graphicx}
\usepackage{booktabs}
\usepackage{subfig}

\usepackage{array}

\begin{document}
\title{Tracking UWB Devices Through Radio Frequency Fingerprinting Is Possible}


\author{\IEEEauthorblockN{Thibaud Ardoin\IEEEauthorrefmark{1},
Niklas Pauli\IEEEauthorrefmark{1},
Benedikt Groß\IEEEauthorrefmark{1}, 
Mahsa Kholghi\IEEEauthorrefmark{2},
Khan Reaz\IEEEauthorrefmark{1}, and
Gerhard Wunder\IEEEauthorrefmark{1},~\IEEEmembership{Fellow,~IEEE}}
\IEEEauthorblockA{\IEEEauthorrefmark{1}  Cybersecurity and AI,
Freie Universität Berlin, and \IEEEauthorrefmark{2} PhySec GmbH, Bochum, Germany}
\IEEEauthorblockA{thibaud.ardoin@fu-berlin.de}
}

\IEEEtitleabstractindextext{
\begin{abstract}

    Ultra-wideband (UWB) is a state-of-the-art technology designed for applications requiring centimeter-level localisation. Its widespread adoption by smartphone manufacturer naturally raises security and privacy concerns. Successfully implementing Radio Frequency Fingerprinting (RFF) to UWB could enable physical layer security, but might also allow undesired tracking of the devices.  
    The scope of this paper is to explore the feasibility of applying RFF to UWB and investigates how well this technique generalizes across different environments. We collected a realistic dataset using off-the-shelf UWB devices with controlled variation in device positioning. Moreover, we developed an improved deep learning pipeline to extract the hardware signature from the signal data. 
    In stable conditions, the extracted RFF achieves over 99\% accuracy. While the accuracy decreases in more changing environments, we still obtain up to 76\% accuracy in untrained locations.




\end{abstract}

}

\maketitle
\IEEEdisplaynontitleabstractindextext

\section{Introduction}

    The Ultra-Wideband (UWB) technology is the current standard for wireless high-resolution and short-range localisation enabling data transmission at high rate. It is therefore the main candidate for smart-city applications that require a precise indoor localisation of the user.
    Indeed, UWB enables a localisation of a \textit{client} in the network by a precision within centimeters. Paired with cheaper and smaller hardware, it has been widely adopted by manufacturer of telecommunication equipment and is present in the latest generations of smartphones \cite{Zheng2023}.
    An example of UWB use case is aiding hospital staff in navigating facilities. With precise localization technology, individuals can open doors or cabinets hands-free and generate reports more efficiently based on the specific context of the room they are in.
    
    Alongside the development of UWB, research on Radio Frequency Fingerprinting (RFF) has recently gained increased attention. It is a type of signal intelligence applied directly on the radio frequency domain. It defines techniques that extract a unique hardware signature for the device that emit the signal. Such a fingerprint is unintentionally introduced by slight variation in the production process of the different physical components. 
    Without altering the quality of the transmitted data, this results in slight changes in the form of the signal. For more precision we define the fingerprint by the following 3 points~\cite{Xie2023}:
    \begin{itemize}
        \item \emph{Differentiable}: Each device is distinguished by a distinctive fingerprint that is discernible from those of other devices.
        \item \emph{Relative stability}: The unique feature should remain as stable as possible over time, despite environmental changes.
        \item \emph{Hardware}: The hardware's condition is the only independent source of the fingerprint. Any other impact on the waveform, such as interference, temperature, time, position, orientation, or software is considered a bias.
    \end{itemize}

    Once a RFF signature is extracted from the signal emitted by a device, it can be used to enhance the security of a network. Since this signature is based solely on the device's hardware, any replay attempt by a malicious third party would alter it. Additionally, masking the signature with software alone would be difficult, as it is derived from the raw signal shape and not from the content of the communication. This forms the foundation of the Physical Layer Security, that aims to enhance device authentication \cite{Jagannath2022}.
    However, this signature can also be employed to track devices without the user's consent. Similarly, as with facial recognition, the unintentionally disclosed features can be employed to track and re-identify a person's device in a variety of environments. In the case of device fingerprinting on the raw communication, it is not necessary to decrypt the data; only signal sniffing is required. 
    
    \subsection{Related work}
    The field of RFF is attracting increasing attention as it becomes evident that such a signature can be extracted and utilised for security purposes. The majority of studies have demonstrated the successful classification of devices across diverse wireless domains, including Wi-Fi, 5G, and Bluetooth. These findings have been synthesised in recent surveys~\cite{Soltanieh2020, Jagannath2022, Xie2023}. 
    The research has explored different methods, with the initial focus being on the mathematical modeling of signal fingerprints. These models aim to leverage prior knowledge about the physical characteristics of the signals for the purposes of RFF extraction.~\cite{Hall2003DETECTIONOT, Ali2017, Tzur2015, Barzegar2020}. Alongside these, classical machine learning techniques, such as Principal Component Analysis (PCA) or t-distributed Stochastic Neighbor Embedding (t-SNE), have been used to classify signals, mostly based on hand-crafted features~\cite{Padilla2014, Danev2009, Yuan2014SpecificEI, UrRehman2012, Ashush2023, Baldini2017}. More recently, deep learning methods, particularly Convolutional Neural Networks (CNNs) applied to 2D spectrograms and 1D convolutions on raw signals, have shown strong performance in signal classification tasks~\cite{Baldini2023, Shao2024, AlShawabka2020, Ding2018, Hanna2021, Saeif2023, Kulhandjian2023}.

    However, a deep dive into these works reveals certain limitations on the evidences of generalization of most of the methods \cite{Alhazbi2023}. 
    Since signal data is not human-readable, it is challenging to identify biases that might lead a machine learning model to classify signals based on factors unrelated to the hardware characteristics.
    Many methods achieve high accuracy in classifying signals based on their emitting devices. However, the typical approach of splitting the data between train and test randomly results in very similar distributions for both sets
    \cite{Ashush2023, Kulhandjian2023}. 
    Signal data can be susceptible to various external biases, both known and unknown. Therefore, it is essential to conduct controlled experiments to rigorously evaluate the model's capacity to generalize across different distributions and quantify its performance under varying conditions.
    With the maturation of RFF research and the adoption of best practices in data handling, recent studies have begun to examine the robustness of the models under varying conditions. 
    For example, a \mbox{Wi-Fi} RFF classification model can experience a 50\% drop in accuracy when the test set is recorded on a different day then the training set \cite{Saeif2023}. In the case of a channel changes, the accuracy can decrease from 85\% to 9\% \cite{AlShawabka2020}. 

    \subsection{Contributions}
    To the best of our knowledge, no research has yet been conducted for RFF on UWB signals, and we would like to close that gap.
    There are two technical characteristics of UWB that could cause greater difficulties to extract a device fingerprint:
    Firstly, the UWB communication is done via short pulse signals. This short duty cycles gives less features from which to perform RFF detection compared to continuous-type wireless protocols.
    Secondly, the key advantage of UWB for end applications is its positional sensitivity. This characteristic results in significant variations in the signal when the position or the surrounding physical environment changes. 
    These substantial changes can potentially hinder the performances of learning model, making it challenging to achieve accurate detection in untrained positions. 
     Considering that positional sensitivity of UWB, we focus on evaluating the robustness of RFF detection under variation of the device position. We conducted experiments in controlled scenarios, ensuring that any environmental variables are either explicitly learned by the model or kept constant throughout the dataset to prevent bias. 
     To conduct a realistic dataset and study, we use real world signal data from off-the-shelf UWB boards. Contrary to most works, we also propose an open-set approach that enables the identification of devices without the need for retraining on them. The method used are data-driven, inspired by state-of-the-art methods in computer vision. 
    The contributions can be summarized in three points:
   \begin{enumerate}
        \item A pioneering investigation of the challenges associated with the extraction of device fingerprint features from UWB signals.
        \item A realistic dataset with controlled variation for different grades of generalisation of our problem \footnotemark.
        \item An improved deep learning pipeline of different method with empirical evidences of their efficiency \footnotemark. 
    \end{enumerate}

    \footnotetext{\url{https://zenodo.org/records/11083153}}
    \footnotetext{\url{https://github.com/Thibaud-Ardoin/UWB-fingerprint}}

    The paper is organized as follows: Sec.~\ref{sec: Data} provides the details about the nature of our data and the protocol to generate it. Sec.~\ref{sec: approach} provides a description of our deep learning-based RFF detection methods, Sec.~\ref{sec: experiments} discusses the results and present further improvements, finally, Sec.~\ref{sec: conclusion} concludes and outlines future work.

\section{Experimental setup and Dataset} \label{sec: Data}

    To ensure fair experiments with controlled variables, we limited all variations to the specific point of study: the location of the emitter. We conducted a data collection campaign that allowed us to create a dataset with scenarios covering varying degrees of difficulty. All other setup variables were intentionally kept constant. This approach represents the most challenging detection scenario, as no additional biases are present to aid detection. Throughout the experiments, we consistently used the same boards, channel, and board orientation.

    \begin{figure}[h]
        \includegraphics[trim={2em 1em 1.5em 4em},clip, width=0.5\textwidth]{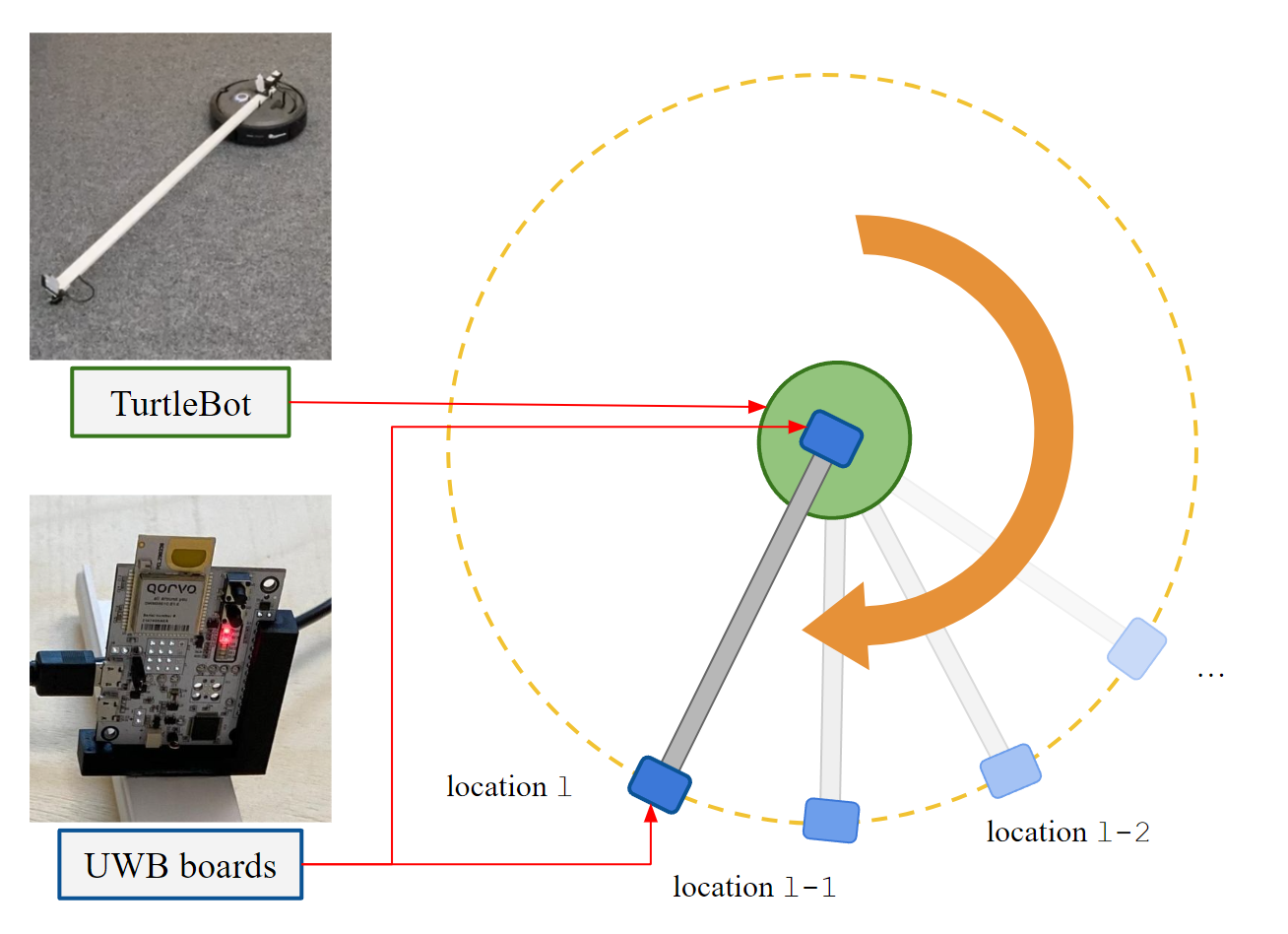}
        \caption{Experimental setup. UWB boards clipped in the 3D-printed mount rotated by the TurtleBot}
        \label{fig: setup1}
    \end{figure}  

    \begin{figure*}[ht!]
        \centering
        \includegraphics[width=0.95\textwidth]{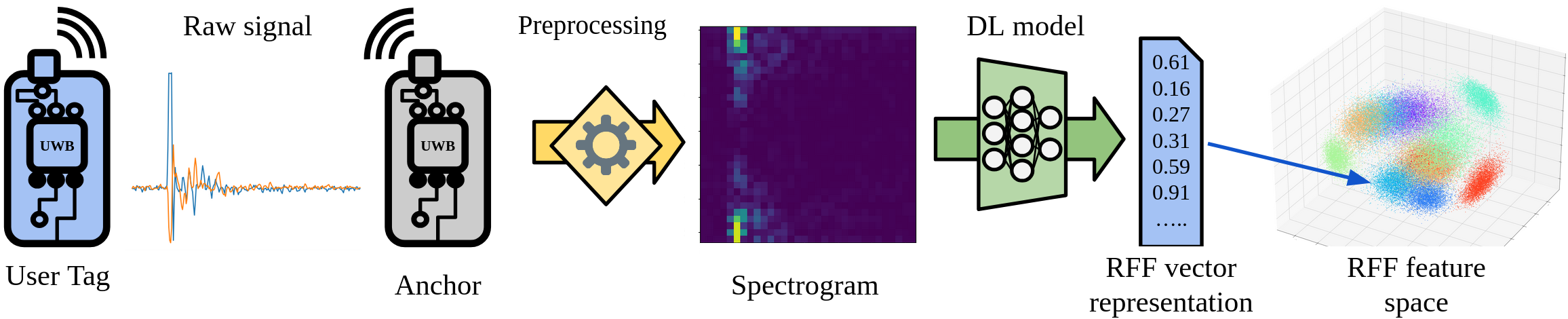}
        \caption{Pipeline of the RFF extraction system through representation learning}
        \label{fig:pipeline}
    \end{figure*}

    \vspace{-1em}
    
    \subsection{Setup} \label{sec: setup}

    In our setup, we use 14 identical development boards from Qorvo, with reference: \textit{DWM3001CDK}. It uses the UWB standard HRP. We use channel 5 for our signals, such that the center frequency is 6489.6 MHz, and the bandwidth 500 MHz.
    One device functions as the receiver throughout the entire data generation process, it represent an anchor of the network. Meanwhile the remaining 13 boards are emitter and serves as user tags.
    We refer to a \textit{measurement} as a complex discrete signal extracted from the UWB pulse emitted by a transmitter and captured by our receiver.
    More precisely, this signal is read from the Channel Impulse Response (CIR) register of the chip, with help of the the off-the-shelf firmware.
    Those measurements represent both location-specific information as well as device-dependent fingerprint information, reflecting the unique hardware imperfection of the transmitter.
    The variations in the UWB chips' control parameters will be limited to the location of the boards within the room and the identities of the devices.
    Consequently, each measurement will be associated with a two-dimensional label composed of \textit{Device ID} and \textit{Device Location}.
    The high sensitivity of UWB to position enables precision at the centimeter level. Therefore, our defined locations must be accurate at a comparable level of precision. Failing to do so may lead the learning model to differentiate devices based on the inexact location rather than on inherent hardware characteristics.
    To achieve this precision, we have designed a 3d printed mount onto which the boards can be clipped, as depicted in Fig.~\ref{fig: setup1}. This mount is attached on a rigid rail, ensuring that the relative positions of the devices remain fixed through the whole data gathering protocol—they consistently face each other at a predetermined distance.
    To vary the device locations within the environment, we rotated the rail holding the devices around the receiver, as illustrated in Fig. \ref{fig: setup1}. To ensure consistent and controlled rotations for each device, we employed a programmable robot, \textit{TurtleBot}. This robot rotates on itself between each series of measurements, dividing the 360° circle into 50 locations for each device. At each precise location, we conducted 2000 measurements.
    We conducted these measurements on two different days and in different positions within the same room. Once at a relative distance of 1 meter, and then again at 2 meters. Our campaign concluded with over 1.5 million measurements, published in open access \cite{AnonDataset}.

    \subsection{Signal pre-processing} \label{sec: signal}



    
    To prevent bias caused by detecting the device solely based on their signal power or latency, we perform a min-max normalisation of the amplitude of the measurements. Additionally, the data is centered so that the peak of the primary pulse occurs at the same time step for every measurement.
    A standard approach to analyzing time-series signals is to transform them into the time-frequency domain for a richer representation. This is here done by applying a discrete Short-Time Fourier Transform (STFT) to the measurements.
    With $X[t,\omega]$ the spectrogram at time index $t$ and frequency index $\omega$, with frequency bins from $0$ to $N$, $x[n]$ the $n$-th element of the measurement in the time domain, $w[n]$ is a \textit{Hann window} function applied to each segment of the input signal with a hop size of $R$, we have:
    \begin{equation}
        X[t, \omega] = \sum_{n=0}^{N-1} x[n] \cdot w[n-tR] \cdot e^{-j\omega n}
    \end{equation}

    \subsection{Dataset} \label{sec: dataset}

    We define a \textit{dataset} as a collection of measurements recorded in the experimental setup described above. Concretely, each data point is a vector of 250 complex values.
    Our learning models operate with two types of datasets: first, the \textit{training set} that will help optimize the parameters of the model through supervised learning. Once trained, the parameter of the model remain fixed and it can be used for inference, as it will be used in practice in production. Secondly, the \textit{test set} is used to evaluate the model. 
    A well-trained model is expected to perform reliably on data with similar variations as its training set, but its performance is uncertain when faced with out-of-distribution data. 
    To evaluate the model's generalization to various realistic applications, we split the data into different scenarios. These scenarios are illustrated in Fig.~\ref{fig:scenarios}:


    \begin{figure}[h!]
        \centering
        \includegraphics[width=0.46\textwidth]{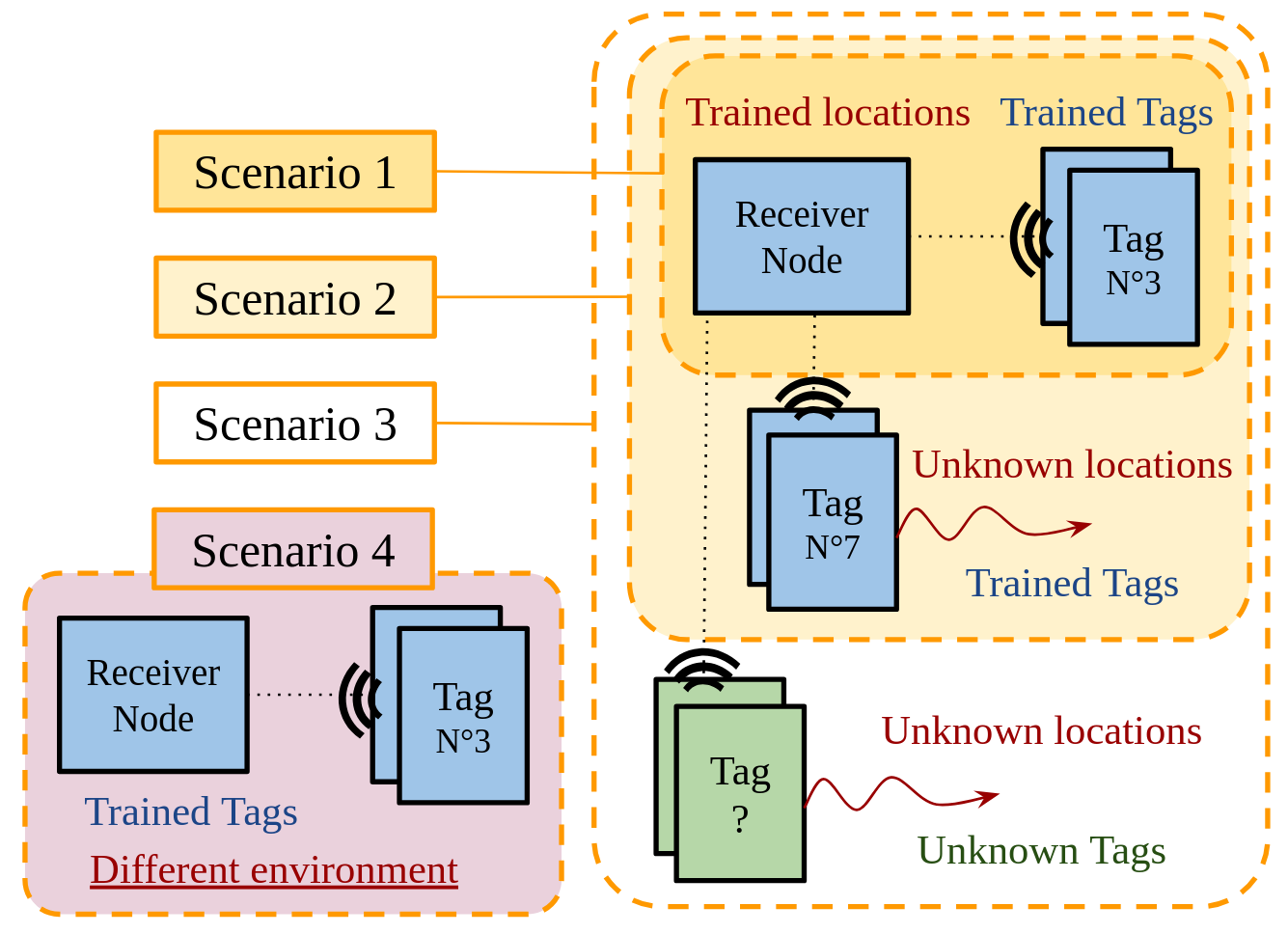}
        \caption{The evaluation scenarios with their degree of complexity}
        \label{fig:scenarios}
    \end{figure}
    
    \begin{itemize}
        \item \emph{Scenario 1: } The fingerprint authentication is conducted between two fixed devices within the UWB network. 
        The test set includes data points with locations and identity labels that were observed during the training procedure.
    
        \item \emph{Scenario 2: } The fingerprint authentication is performed between a network \emph{anchor} and a \emph{tag} moving around the environment. 
        The feature extraction is trained specifically for the tag's ID, excluding data from three specified test locations.
    
        \item \emph{Scenario 3: } The fingerprint authentication is carried out between an anchor and an unknown tag in an unknown position. This is an open-set scenario, the network generates a trusted feature embedding of that tag to serve as reference representation. Future feature embedding can then be compared to this reference to re-identify the tag. Three device IDs and three locations are excluded of the training set and evaluated on.

        \item \emph{Scenario 4: } The RFF authentication is done in a significantly different environment from where the model was trained. The test set consists of all the measurements done with a relative distance of 2 meters between transmitter and receiver. These measurements were taken in a different position within the room and on a different day compared to the training data.

    \end{itemize}


\section{Approach} \label{sec: approach}

    To explore the feasibility of extracting RFF information from UWB devices, we adopt a data-driven approach. We proposed an enhanced Deep Learning (DL) architecture specifically designed to address open-set challenges, such as device re-identification in dynamic environments.

    \subsection{Re-identification system in Open Set problem}

\begin{figure}[h]
          \centering
          \footnotesize
          \subfloat[Raw data]{\includegraphics[width=0.23\textwidth]{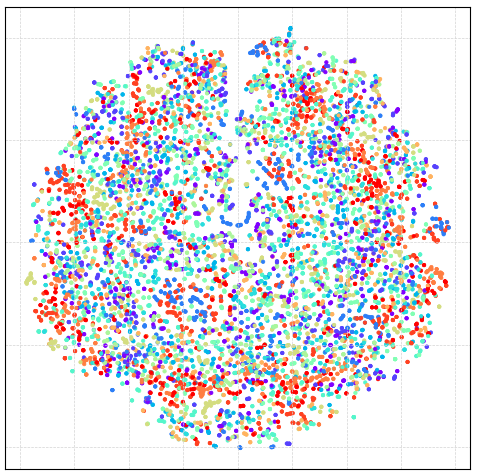}}
          \hfill
          \subfloat[Structured by model]{\includegraphics[width=0.23\textwidth]{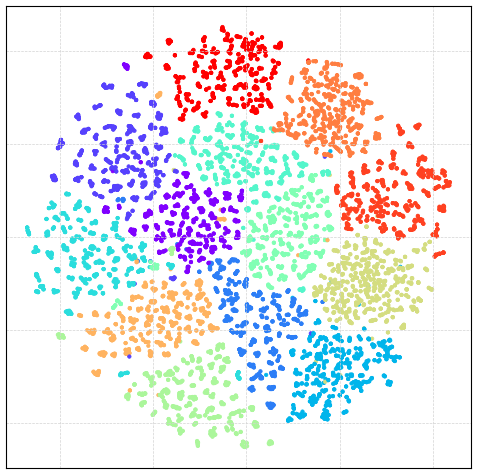}}
          \caption{t-SNE visualisation of the data before and after the projection of the DL model in the RFF feature space.}
          \label{fig: t-sne visulisation}
        \end{figure}

        In order to be closer to real case scenarios, we chose to build our architecture as a re-identification system~\cite{Hanna2021}. Such system differs from a simple classification task because, instead of just assigning a label to input data, it focuses on learning a meaningful representation of the data. The model converts the input signal measurement into a feature vector, ensuring that features from different measurements of the same device ID are placed closely together, while features from different IDs are kept far apart.
        Once the model is trained, the system can store known identities as reference points in this feature space. Subsequently, when new input signal is received, it is projected into the feature space and compared to these references. If it is close to a stored reference point, the corresponding identity is assigned; if not, it is treated as a new identity. 
        This method effectively addresses the open-set problem by handling both known and unknown identities.
        The goal of training the model is to achieve an optimal representation of device features, independent of positional changes, enabling accurate and reliable re-identification. The t-SNE visualisation in Fig.~\ref{fig: t-sne visulisation} illustrates how our trained model structures the data based on device identities, depicted in different colors.
    
        \captionsetup[subfigure]{labelformat=empty}

\subsection{Deep Learning models} \label{sec: deep learning}

      \begin{figure}[hb]
        \centering
        \includegraphics[width=0.20\textwidth]{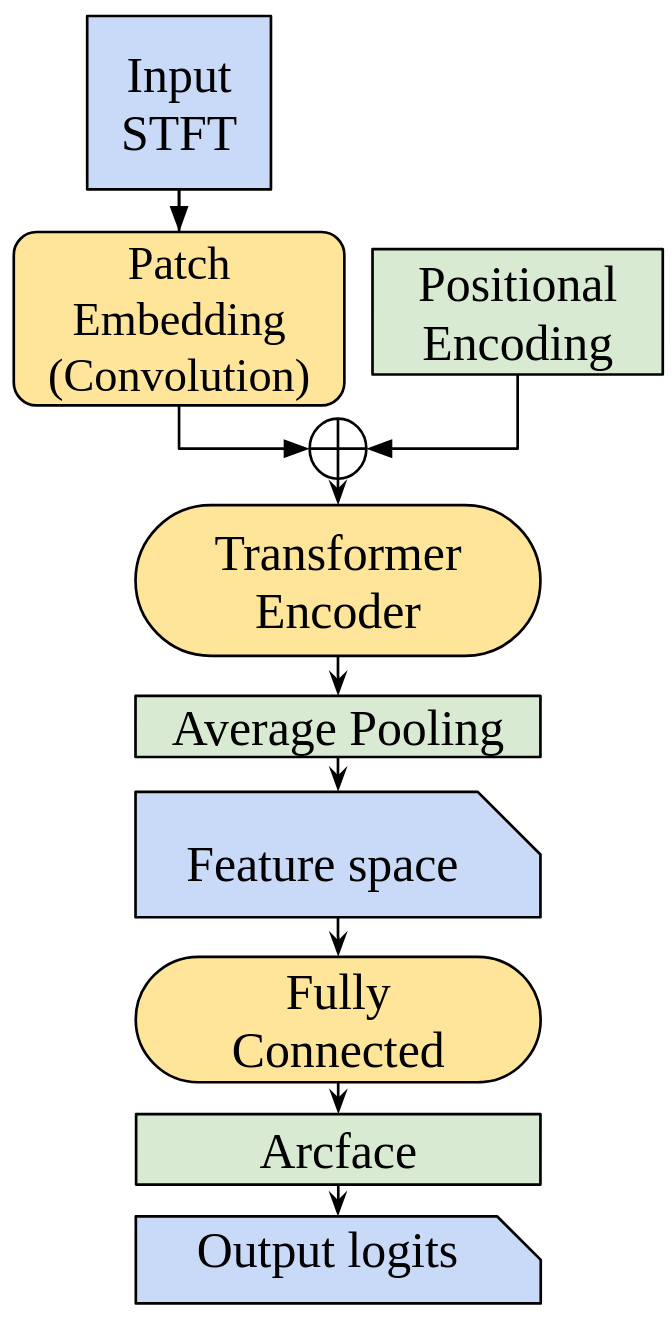}
        \qquad
        \begin{tabular}[b]{cc}
        \hline
        \textbf{Input} & $32 \times 32$ \\ \hline
        
        \textbf{Patch} \\ \hline
          Kernel size & 8 \\
          Stride & 8 \\
          Features & $ 6 \times 32$ \\
          \hline 
        \textbf{Transformer} \\ \hline
          Layers & 1 \\
          Heads & 6 \\
          Hidden size & 32 \\
          Hidden  & 32 \\ 
          \hline
        \textbf{Latent space} & 192 \\ \hline
        \textbf{Arcface} \\ \hline
            Margin & 0.1 \\ 
            Scale & 64 \\ 
        \hline
        \vspace{0em}
        \end{tabular}
        \caption{ViT architecture with parameters}
        \label{fig: ViT-architecture}
      \end{figure}

\newcolumntype{L}[1]{>{\raggedright\arraybackslash}p{#1}}
  \begin{table*}[!t]
      \caption{Evaluation of the model representations, compared to the metric performances of a random projection}
        \centering
        \begin{tabular}{@{}p{0.10\textwidth}*{12}{L{\dimexpr0.1\textwidth-5\tabcolsep\relax}}@{}}
        & \multicolumn{3}{c}{CNN} &
        \multicolumn{1}{c}{} & 
        \multicolumn{3}{c}{ViT} & 
        \multicolumn{1}{c}{} & 
        \multicolumn{3}{c}{Random projection} &
        \multicolumn{1}{c}{} \\
        \cmidrule(r{4pt}){2-4} \cmidrule(l){6-8} \cmidrule(l){10-12}
        & CF1 & CMC & AUROC & 
        & CF1 & CMC & AUROC & 
        & CF1 & CMC & AUROC \\
        \midrule
        Scenario 1 & $96.4\%$ & $86.1\%$ & $0.93$ & 
        & $99.9\%$ & $99.7\%$ & $0.99$ & 
        & $7.7\%$ & $7.7\%$ & $0.5$ \\
        
        Scenario 2 & $61.5\%$ & $52.8\%$ & $0.89$ & 
         & $64.6\%$ & $53.4\%$ & $0.92$ & 
        & -- & -- & -- \\
        
        Scenario 3 & $14.0\%$ & $16.9\%$ & $0.63$ & 
        & $34.9\%$ & $18.9\%$ & $0.76$ &
        & -- & -- & -- \\
        
        Scenario 4 & $11.2\%$ & $13.1\%$ & $0.62$ & 
        & $14.6\%$ & $13.4\%$ & $0.64$ & 
        & -- & -- & --  \\
        
        \bottomrule
        \end{tabular}
        \label{table:ML results}
\end{table*}

    For the study of the generalization of RFF, we propose two detailed DL architectures. As a comparison baseline, we use a CNN architecture that is comparable to the architectures employed in the recent RFF literature~\cite{Ding2018, AlShawabka2020}.
    In addition, we compare the CNN baseline to a model inspired by the Vision Transformer (ViT) \cite{Dosovitskiy2021}, with the implemented architecture visualized in Fig.~\ref{fig: ViT-architecture}. This choice is motivated by the superior performance of transformer-based architectures compared to CNNs across various domains \cite{Lin2021}. To the best of our knowledge this transformer-based architecture has not yet been applied to the RFF detection domain.

    Our architectural and parameter choices are guided by an empirical study of the pipeline and the hyper-parameters have been optimised through grid search. 
    For better comparison purpose, the two models share a comparable architecture, including multiple layers interspersed by Rectified Linear Units (ReLU) activation functions to introduce non-linearity. Training utilises the \textit{Adam} optimizer with a learning rate of $10^{-4}$ and a batch size of $512$, over $50$ epochs.
    The training objective is a cross-entropy classification of the device ID, given by a linear projection of the feature space performed by a final fully connected layer. We chose not to use the triplet loss or other contrastive losses because they are less stable and require more hyper-parameter fine-tuning. But note that they also provide good results.
    For the ViT, we incorporate the ArcFace loss \cite{Deng2018}, a standard element of the state-of-the-art facial recognition models. This additive angular margin loss enhance the intra-class compactness and inter-class separation. It is especially relevant for open-set classification to better distinguish new identities.       
    With  $\theta_{y_i}$ the angle between the feature vector and the class weight vector, $s$ the scaling factor that adjusts the softmax output and $m$ the angular margin that enhances class separability, we have:
    \begin{equation}
        L_{AF} = -\frac{1}{N} \sum_{i=1}^N \log \frac{e^{s \cdot \cos(\theta_{y_i} + m)}}{e^{s \cdot \cos(\theta_{y_i} + m)} + \sum_{j = 1,j \neq y_i}^n e^{s \cdot \cos(\theta_j)}}
    \end{equation}


\section{experimental results} \label{sec: experiments}

    The results of the RFF recognition by our models are presented in Table \ref{table:ML results} with metrics that provide various insights into the model's performances.
    As expected, our results vary according to the difficulty of the scenario. Nevertheless, we can provide hints of improvements with additional experiments and thus confirm the existence of a clear detection of a RFF information.

    \subsection{Metrics} \label{sec: metric}

        Like many related works, we initially assess classification accuracy in a close-set problem. More precisely, we use the \textit{F1-score} of the classification (CF1), which is the harmonic mean of precision and recall.   
        This accuracy comes from a shallow linear classifier trained on top of the feature representation of our DL model. 
        While this metric demonstrates the model's ability to create separable representations, it's important to note that separability alone is not the sole desired outcome.

        Indeed, in a real-world scenario the model is not retrained each time a new ID joins the system, it is an open-set problem as described in \textit{scenario 3}. 
        To address this, we will introduce a metric widely used in the field of biometrics: the Cumulative Matching Characteristics (CMC) \cite{Gray2007EvaluatingAM}. This metric measures the probability of finding a correct match for a given query identity within the top-N ranked matches. Formally, the CMC curve is defined as follows:
        \begin{equation}
            CMC(N) = \frac{1}{M} \sum_{i=1}^{M} P_i(N)
        \end{equation}

        With $M$ the total number of queries and $P_i(N)$ the empirical probability that the correct match for the $i$-th query is among the nearest $N$ elements of the reference representation gallery. For clarity, we focus on the top-1 ranking, as it provides the more valuable information \cite{Zhao2014}. 
        In our experiments, the reference points are averaged features from the training set. If we are in an open-set situation like \textit{scenario 3} we use one unseen test position to compute average features as reference point. The queries are the rest of the test data.
        
        Additionally, to assess the consistency of the RFF projection on the test set, we evaluate the quality of the aggregation regardless of whether it can be associated with a known ID, as proposed in FaceNet \cite{Schroff2015}.
        For this purpose, we use the Receiver operating characteristic (ROC) curve. 
        This curve illustrates the trade-off between false-positive rate (FPR) and true-positive rate (TPR) for a given neighborhood size around the data points. With true positive (TP), true negative (TN), false positive (FP), and false negative (FN), we define:
        
        \vspace{0.8em}
        \begin{minipage}{.45\linewidth}
            \begin{equation*}
              TPR = \frac{TP}{TP + FN}, 
            \end{equation*}
        \end{minipage}
        \begin{minipage}{.45\linewidth}
            \begin{equation}
                FPR = \frac{FP}{FP + TN}
            \end{equation}
        \end{minipage}
        \vspace{0.8em}
        
        To summarise the ROC curve in a single interpretable value, we compute the the Area Under the ROC curve (AUROC).

        \captionsetup[subfigure]{labelformat=empty}
        \begin{figure}[b]
          \centering
          \footnotesize
          \subfloat[CNN]{\includegraphics[trim={5em 0 14em 0},clip, width=0.23\textwidth]{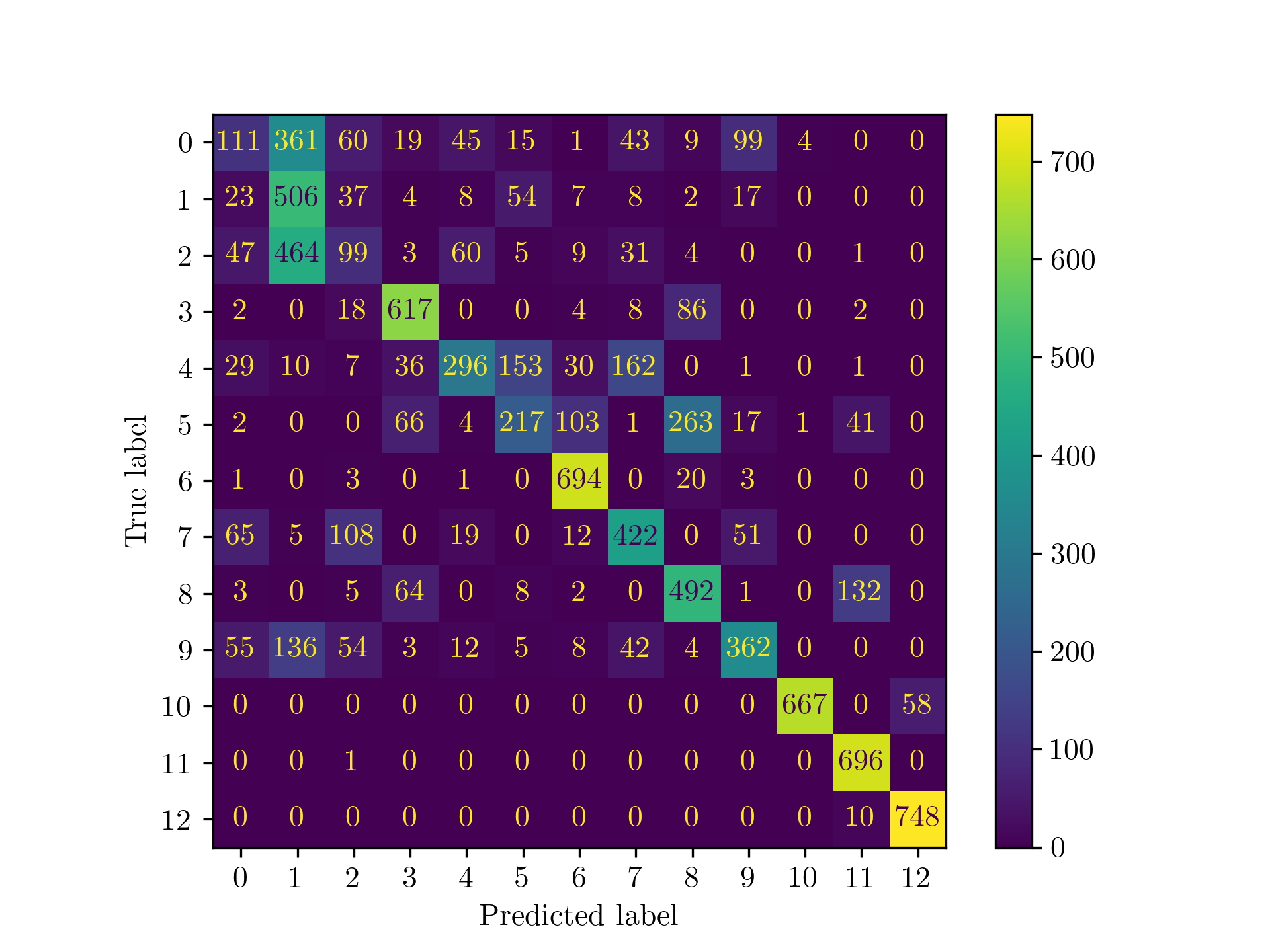}}
          \hfill
          \subfloat[ViT]{\includegraphics[trim={5em 0 14em 0},clip, width=0.23\textwidth]{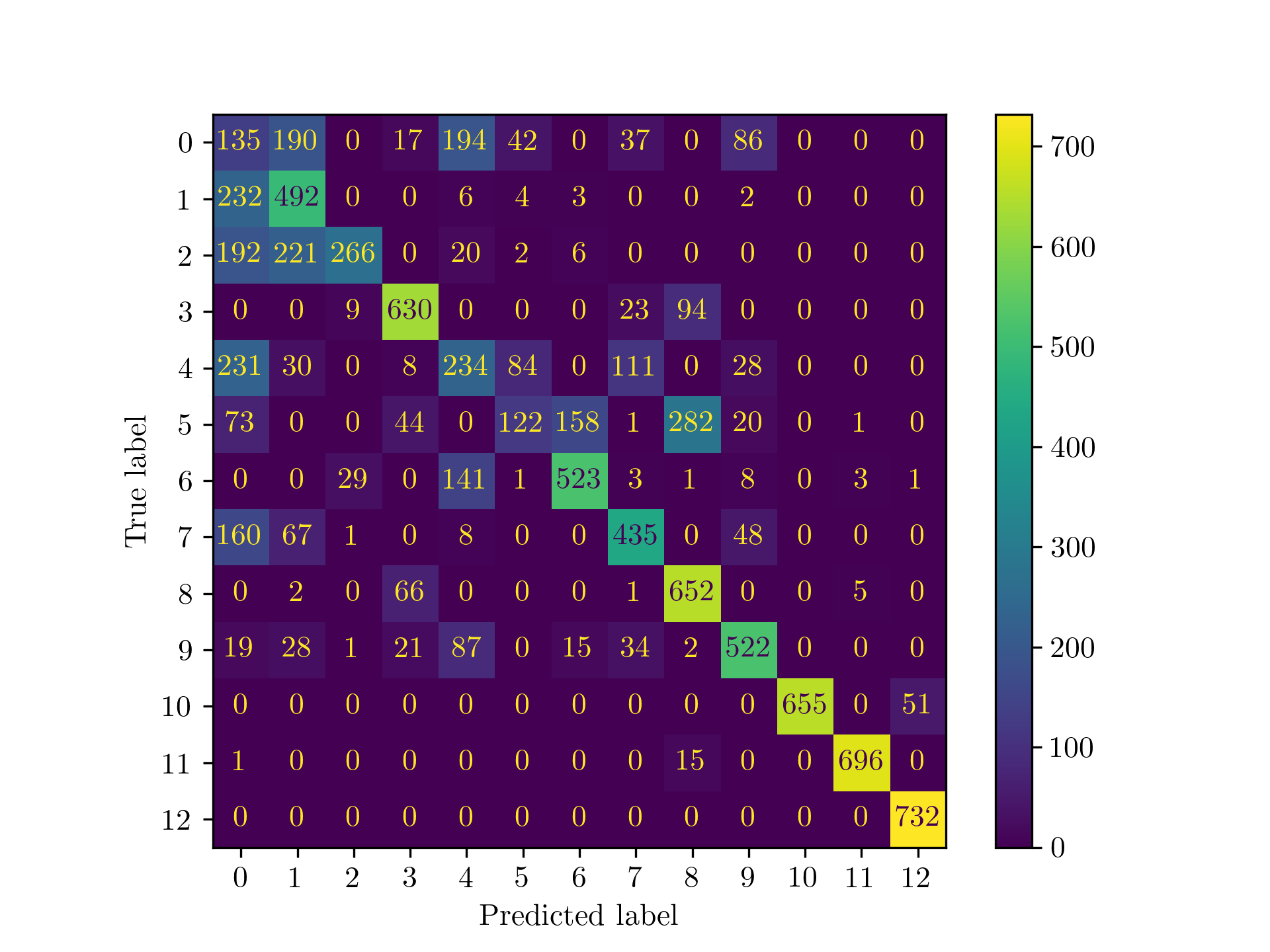}}
          \caption{Confusion matrix on 9300 samples of scenario 2 test set, for two distinct architectures.}
          \label{fig: confusion-matrix}
        \end{figure}     
    
        \subsection{Discussions} \label{sec: evaluation}

        Firstly, the near perfect accuracy in \textit{scenario 1} demonstrate the reliability of RFF for the specific case of fixed communicating UWB nodes in a network.
        The introduction of controlled variations in the test results as expected in a decline in the model's performance. However, the fingerprint features remain recoverable, and the feature space consistent with above $0.9$ AUROC. The confusion matrix in Fig.~\ref{fig: confusion-matrix} indicates that different models exhibit the same difficulty to distinguishes specific devices. Apart from reducing the overall performance, this could hint toward the presence of more similar hardware signatures among certain devices.

        For the open-set problem of \textit{scenario 3}, we excluded 3 devices from the training process and we re-identify them among all the reference points of the 13 devices. with a AUROC of $0.76$ it indicates still  strong consistency of the extraction in a completely unknown setup. The limitation does not lie in the RFF extraction itself, but rather in the consistency of that extraction when compared across different locations.
        In the out-of-distribution \textit{scenario 4}, as expected, we reach the limit of the generalisation of the model, as learning algorithms provide no guarantee of functioning on data that variate on dimensions not present in the training set. 
        However, despite the challenges posed by data collected on a different day, with different distance between devices, and in a different location the extraction give accuracy almost twice as good as random. 
        Therefore, we interpret these results as an indication of the potential to generalize detection in unseen environments. Nevertheless, achieving generalized detection would require more diversity in the data.
        For comparison, the first open-source facial recognition datasets, such as CelebA \cite{liu2015faceattributes}, contain around $10,000$ different identities captured in various and uncontrolled environments.

        \begin{table}[h]
            \centering
            \caption{CF1 score on scenario 2 for detection techniques with multiple samples. We have Same location (Sl) or Different Location (Dl) for the different measurements. Used as Concatenated Input (CI) of the ViT or as Voting (V) of individual classification probabilities.}
            \begin{tabular}{lccc}
                \toprule
                & \textbf{1} sample & \textbf{3} samples & \textbf{10} samples \\
                \midrule
                Sl-V & $62.1\%$ & $70.3\%$ & $71.7\%$ \\
                Dl-V & - & $71.7\%$ & $73.3\%$ \\
                Sl-CI & - & $69.8\%$ & - \\
                Dl-CI & - & $75.7\%$ & $76.9\%$ \\
                \bottomrule
            \end{tabular}
            \label{table: multi-samples}
        \end{table}

        Note that the results of Table \ref{table:ML results} represent raw 1-shot performances aimed at providing an initial assessment of the possibilities of RFF on UWB. In practical cases, accuracy could be enhanced by collecting multiple signal samples per device, for identification purposes for example. Hence each sample is sub-nanosecond long, it is realistic for a protocol implementation to be based on the collection of multiple samples. This approach provides the model with more contextual information to extract a valid fingerprint feature, especially if the different samples originate from slightly different device locations.
        To demonstrate the potential enhancement of a multi-sample system, we present results in Table \ref{table: multi-samples}.  
        This model trained on concatenated samples is compared against a more naive voting system that aggregates the prediction probabilities of a single-sample ViT model. These approaches are further distinguished based on whether the samples where collected in the same location or in different test locations. We can see a clear improvement in performance with additional samples, particularly when the model is specifically trained on the concatenated samples of different locations.
        In this comparison, we used a ViT model trained on raw measurements, because it did not require additional hyper-parameter tuning of the STFT to perform on concatenated signals.

        To evaluate the model's robustness against board manipulation, we covered the antenna with hot glue. We recorded the signal at 3 different stages: before manipulation, with the glue and after removal, with 5 different boards. We projected the data in the feature space with our ViT model trained in \textit{scenario 2} on the previous dataset. Training a linear classifier in the feature space on the data before manipulation. This linear classifier gets 99\% CF1 with the data after removal but only 44\% CF1 if the glue is on. We can then conclude that our model is not perfectly robust to such manipulations but still performs significantly better then random guessing.

        \begin{figure}[h!]
            \centering
            \includegraphics[width=0.5\linewidth]{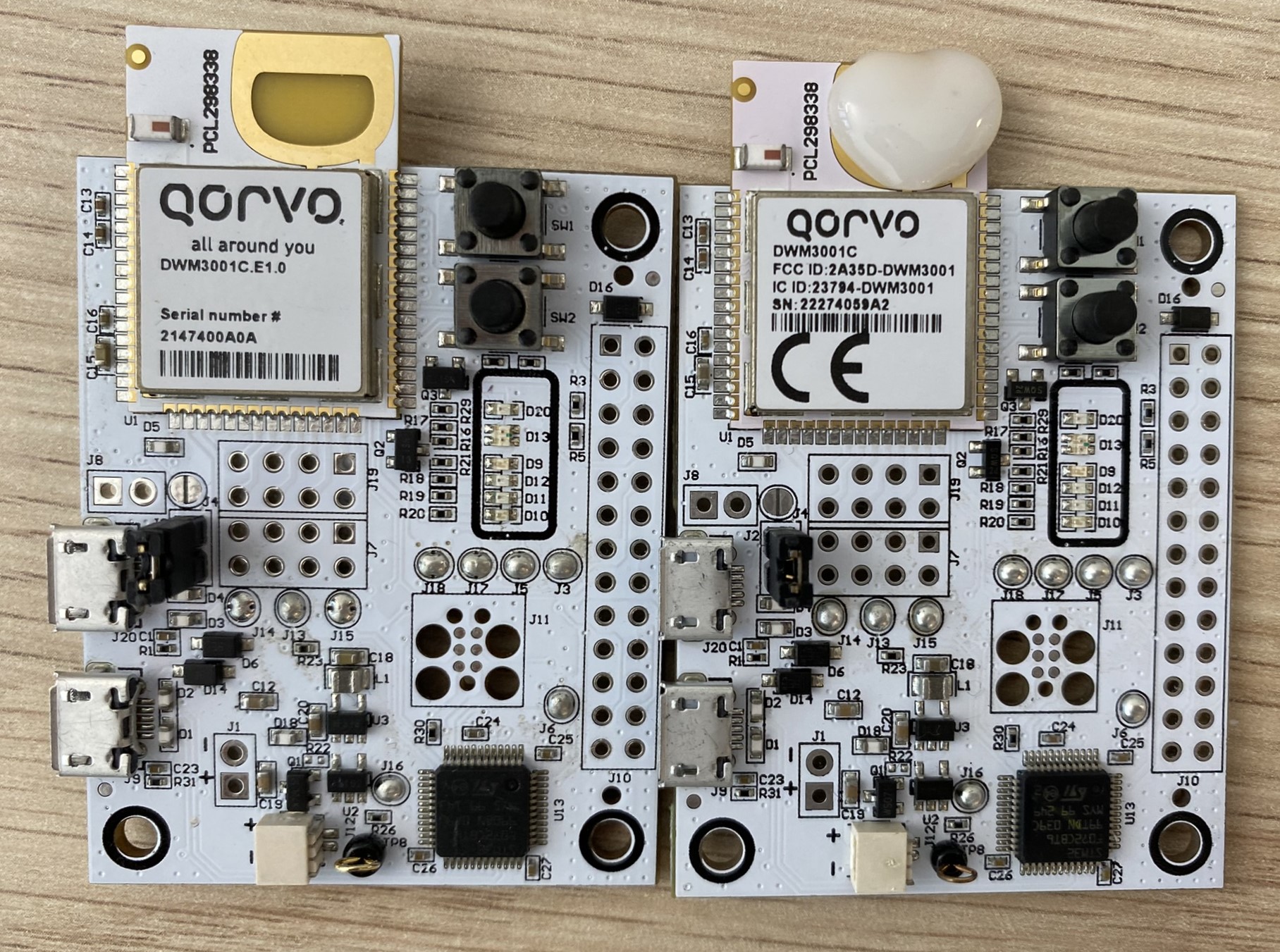}
            \caption{UWB boards with and without glued antenna}
            \label{fig:manipulated-boards}
        \end{figure}

    \subsection{Limitation}

        Our study does not evaluate the network overhead associated with incorporating a RFF based authentication system using DL models. The ViT model, with $200k$ parameters, is computationally intensive for a continuous usage. Therefore it is best suited for low-trust environments where additional authentication can enhance security.
        While API integration is a potential solution, further research is needed to explore extensions through Federated Learning or the use of smaller models.


\section{Conclusion and future work} \label{sec: conclusion}

    The critical aspect of a reliable hardware-based RFF for tracking of UWB devices lies in its ability to generalize the learning process.
    The key insight gleaned from our work is the successful extraction of device fingerprints from raw und unbiased UWB signals. Nevertheless, we have observed that the effectiveness of this extraction is significantly influenced by environmental stability and dataset diversity. Greater diversity in the data would likely lead to improved generalization for extracting fingerprints from new devices and in new environments. Therefore, rigorous data collection procedures and it sharing are essential. Proving the absence of bias in the data is inherently challenging, as interpretation of raw signals is limited. In the future, having access to multiple datasets could enhance the reliability of evaluations by testing across different environments.    
    Our future work will concentrate on assessing the performances of our technique across additional technologies and in diverse environments. Building upon the novel ideas proposed in this study --such as the use of Transformer or Arcface loss-- inspired by data-driven approaches in other domains, we aim to introduce more variance into our datasets to enhance the robustness of our models. Furthermore, we recognize the need to investigate numerous other environmental parameters that may distort the device fingerprint, providing a comprehensive understanding of the challenges involved in real-world implementation.

    \section*{Acknowledgments}
    \noindent TA and GW was also supported by the Federal Ministry of Education and Research of Germany (BMBF) in the programme of “Souverän. Digital. Vernetzt.”, joint project “AIgenCY : Chances und Risks of Generative AI in Cybersecurity”, project identification number 16KIS2013, and by BMBF joint project “6G- RIC: 6G Research and Innovation Cluster”, project identification number 16KISK020K, as well as the BMBF joint project “UltraSec: Security Architecture for UWB-based Application Platform”, project identification number 16KIS1682

\bibliographystyle{IEEEtran}
\bibliography{bib}

\begin{thebibliography}{10}
\providecommand{\url}[1]{#1}
\csname url@samestyle\endcsname
\providecommand{\newblock}{\relax}
\providecommand{\bibinfo}[2]{#2}
\providecommand{\BIBentrySTDinterwordspacing}{\spaceskip=0pt\relax}
\providecommand{\BIBentryALTinterwordstretchfactor}{4}
\providecommand{\BIBentryALTinterwordspacing}{\spaceskip=\fontdimen2\font plus
\BIBentryALTinterwordstretchfactor\fontdimen3\font minus \fontdimen4\font\relax}
\providecommand{\BIBforeignlanguage}[2]{{%
\expandafter\ifx\csname l@#1\endcsname\relax
\typeout{** WARNING: IEEEtran.bst: No hyphenation pattern has been}%
\typeout{** loaded for the language `#1'. Using the pattern for}%
\typeout{** the default language instead.}%
\else
\language=\csname l@#1\endcsname
\fi
#2}}
\providecommand{\BIBdecl}{\relax}
\BIBdecl

\bibitem{Zheng2023}
\BIBentryALTinterwordspacing
C.~Zheng, Y.~Ge, and A.~Guo, ``Ultra-wideband technology: Characteristics, applications and challenges,'' 2023. [Online]. Available: \url{https://arxiv.org/abs/2307.13066}
\BIBentrySTDinterwordspacing

\bibitem{Xie2023}
\BIBentryALTinterwordspacing
L.~Xie, L.~Peng, J.~Zhang, and A.~Hu, ``Radio frequency fingerprint identification for internet of things: A survey,'' \emph{Security and Safety}, vol.~3, p. 2023022, Sep. 2023. [Online]. Available: \url{http://dx.doi.org/10.1051/sands/2023022}
\BIBentrySTDinterwordspacing

\bibitem{Jagannath2022}
\BIBentryALTinterwordspacing
A.~Jagannath, J.~Jagannath, and P.~S. P.~V. Kumar, ``A comprehensive survey on radio frequency (rf) fingerprinting: Traditional approaches, deep learning, and open challenges,'' \emph{Computer Networks}, vol. 219, p. 109455, Dec. 2022. [Online]. Available: \url{http://dx.doi.org/10.1016/j.comnet.2022.109455}
\BIBentrySTDinterwordspacing

\bibitem{Soltanieh2020}
N.~Soltanieh, Y.~Norouzi, Y.~Yang, and N.~C. Karmakar, ``A review of radio frequency fingerprinting techniques,'' \emph{IEEE Journal of Radio Frequency Identification}, vol.~4, no.~3, pp. 222--233, 2020.

\bibitem{Hall2003DETECTIONOT}
\BIBentryALTinterwordspacing
J.~Hall, M.~Barbeau, E.~Kranakis \emph{et~al.}, ``Detection of transient in radio frequency fingerprinting using signal phase,'' in \emph{Wireless and optical communications}, vol.~9, 2003, p.~13. [Online]. Available: \url{https://api.semanticscholar.org/CorpusID:8404182}
\BIBentrySTDinterwordspacing

\bibitem{Ali2017}
\BIBentryALTinterwordspacing
A.~M. Ali, E.~Uzundurukan, and A.~Kara, ``Improvements on transient signal detection for rf fingerprinting,'' in \emph{2017 25th Signal Processing and Communications Applications Conference (SIU)}.\hskip 1em plus 0.5em minus 0.4em\relax IEEE, May 2017. [Online]. Available: \url{http://dx.doi.org/10.1109/SIU.2017.7960417}
\BIBentrySTDinterwordspacing

\bibitem{Tzur2015}
\BIBentryALTinterwordspacing
A.~Tzur, O.~Amrani, and A.~Wool, ``Direction finding of rogue wi-fi access points using an off-the-shelf mimo–ofdm receiver,'' \emph{Physical Communication}, vol.~17, p. 149–164, Dec. 2015. [Online]. Available: \url{http://dx.doi.org/10.1016/j.phycom.2015.08.010}
\BIBentrySTDinterwordspacing

\bibitem{Barzegar2020}
M.~B. Khalilsarai, B.~Gross, S.~Stefanatos, G.~Wunder, and G.~Caire, ``Wifi-based channel impulse response estimation and localization via multi-band splicing,'' in \emph{GLOBECOM 2020 - 2020 IEEE Global Communications Conference}, 2020, pp. 1--6.

\bibitem{Padilla2014}
\BIBentryALTinterwordspacing
J.~L. Padilla, P.~Padilla, J.~F. Valenzuela-Valdés, J.~Ramírez, and J.~M. Górriz, ``Rf fingerprint measurements for the identification of devices in wireless communication networks based on feature reduction and subspace transformation,'' \emph{Measurement}, vol.~58, p. 468–475, Dec. 2014. [Online]. Available: \url{http://dx.doi.org/10.1016/j.measurement.2014.09.009}
\BIBentrySTDinterwordspacing

\bibitem{Danev2009}
B.~Danev and S.~Capkun, ``Transient-based identification of wireless sensor nodes,'' in \emph{2009 International Conference on Information Processing in Sensor Networks}, 2009, pp. 25--36.

\bibitem{Yuan2014SpecificEI}
\BIBentryALTinterwordspacing
Y.~Yuan, Z.~Huang, H.~Wu, and X.~Wang, ``Specific emitter identification based on hilbert-huang transform-based time-frequency-energy distribution features,'' \emph{IET Commun.}, vol.~8, pp. 2404--2412, 2014. [Online]. Available: \url{https://api.semanticscholar.org/CorpusID:23207381}
\BIBentrySTDinterwordspacing

\bibitem{UrRehman2012}
\BIBentryALTinterwordspacing
S.~Ur~Rehman, K.~Sowerby, and C.~Coghill, ``Rf fingerprint extraction from the energy envelope of an instantaneous transient signal,'' in \emph{2012 Australian Communications Theory Workshop (AusCTW)}.\hskip 1em plus 0.5em minus 0.4em\relax IEEE, Jan. 2012. [Online]. Available: \url{http://dx.doi.org/10.1109/AusCTW.2012.6164912}
\BIBentrySTDinterwordspacing

\bibitem{Ashush2023}
\BIBentryALTinterwordspacing
N.~Ashush, S.~Greenberg, E.~Manor, and Y.~Ben-Shimol, ``Unsupervised drones swarm characterization using rf signals analysis and machine learning methods,'' \emph{Sensors}, vol.~23, no.~3, p. 1589, Feb. 2023. [Online]. Available: \url{http://dx.doi.org/10.3390/s23031589}
\BIBentrySTDinterwordspacing

\bibitem{Baldini2017}
\BIBentryALTinterwordspacing
G.~Baldini, R.~Giuliani, G.~Steri, and R.~Neisse, ``Physical layer authentication of internet of things wireless devices through permutation and dispersion entropy,'' in \emph{2017 Global Internet of Things Summit (GIoTS)}.\hskip 1em plus 0.5em minus 0.4em\relax IEEE, Jun. 2017. [Online]. Available: \url{http://dx.doi.org/10.1109/GIOTS.2017.8016272}
\BIBentrySTDinterwordspacing

\bibitem{Baldini2023}
\BIBentryALTinterwordspacing
G.~Baldini, ``Transient‐based radio frequency fingerprinting with adaptive ensemble of transforms and convolutional neural network,'' \emph{Electronics Letters}, vol.~59, no.~22, Nov. 2023. [Online]. Available: \url{http://dx.doi.org/10.1049/ell2.13032}
\BIBentrySTDinterwordspacing

\bibitem{Shao2024}
\BIBentryALTinterwordspacing
Y.~Shao, J.~Liu, Y.~Zeng, and Y.~Gong, ``A radio frequency fingerprinting scheme using learnable signal representation,'' \emph{IEEE Communications Letters}, vol.~28, no.~1, p. 73–77, Jan. 2024. [Online]. Available: \url{http://dx.doi.org/10.1109/LCOMM.2023.3336901}
\BIBentrySTDinterwordspacing

\bibitem{AlShawabka2020}
\BIBentryALTinterwordspacing
A.~Al-Shawabka, F.~Restuccia, S.~D’Oro, T.~Jian, B.~Costa~Rendon, N.~Soltani, J.~Dy, S.~Ioannidis, K.~Chowdhury, and T.~Melodia, ``Exposing the fingerprint: Dissecting the impact of the wireless channel on radio fingerprinting,'' in \emph{IEEE INFOCOM 2020 - IEEE Conference on Computer Communications}.\hskip 1em plus 0.5em minus 0.4em\relax IEEE, Jul. 2020. [Online]. Available: \url{http://dx.doi.org/10.1109/INFOCOM41043.2020.9155259}
\BIBentrySTDinterwordspacing

\bibitem{Ding2018}
\BIBentryALTinterwordspacing
L.~Ding, S.~Wang, F.~Wang, and W.~Zhang, ``Specific emitter identification via convolutional neural networks,'' \emph{IEEE Communications Letters}, vol.~22, no.~12, p. 2591–2594, Dec. 2018. [Online]. Available: \url{http://dx.doi.org/10.1109/LCOMM.2018.2871465}
\BIBentrySTDinterwordspacing

\bibitem{Hanna2021}
\BIBentryALTinterwordspacing
S.~Hanna, S.~Karunaratne, and D.~Cabric, ``Open set wireless transmitter authorization: Deep learning approaches and dataset considerations,'' \emph{IEEE Transactions on Cognitive Communications and Networking}, vol.~7, no.~1, p. 59–72, Mar. 2021. [Online]. Available: \url{http://dx.doi.org/10.1109/TCCN.2020.3043332}
\BIBentrySTDinterwordspacing

\bibitem{Saeif2023}
\BIBentryALTinterwordspacing
A.~Saeif, S.~Savio, and O.~Gabriele, ``The day-after-tomorrow: On the performance of radio fingerprinting over time,'' in \emph{Annual Computer Security Applications Conference}, ser. ACSAC ’23.\hskip 1em plus 0.5em minus 0.4em\relax ACM, Dec. 2023. [Online]. Available: \url{http://dx.doi.org/10.1145/3627106.3627192}
\BIBentrySTDinterwordspacing

\bibitem{Kulhandjian2023}
\BIBentryALTinterwordspacing
H.~Kulhandjian, E.~Batz, E.~Garcia, S.~Vega, S.~Velma, M.~Kulhandjian, C.~D’Amours, B.~Kantarci, and T.~Mukherjee, ``Ai-based rf-fingerprinting framework and implementation using software-defined radios,'' in \emph{2023 International Conference on Computing, Networking and Communications (ICNC)}.\hskip 1em plus 0.5em minus 0.4em\relax IEEE, Feb. 2023. [Online]. Available: \url{http://dx.doi.org/10.1109/ICNC57223.2023.10074023}
\BIBentrySTDinterwordspacing

\bibitem{Alhazbi2023}
\BIBentryALTinterwordspacing
S.~Alhazbi, A.~Hussain, S.~Sciancalepore, G.~Oligeri, and P.~Papadimitratos, ``Challenges of radio frequency fingerprinting: From data collection to deployment,'' 2023. [Online]. Available: \url{https://arxiv.org/abs/2310.16406}
\BIBentrySTDinterwordspacing

\bibitem{AnonDataset}
\BIBentryALTinterwordspacing
T.~Ardoin and M.~Kholghi, ``Ruff -- rotating uwb for fingerprint,'' 2024. [Online]. Available: \url{https://zenodo.org/doi/10.5281/zenodo.11083153}
\BIBentrySTDinterwordspacing

\bibitem{Dosovitskiy2021}
\BIBentryALTinterwordspacing
A.~Dosovitskiy, L.~Beyer, A.~Kolesnikov, D.~Weissenborn, X.~Zhai, T.~Unterthiner, M.~Dehghani, M.~Minderer, G.~Heigold, S.~Gelly, J.~Uszkoreit, and N.~Houlsby, ``An image is worth 16x16 words: Transformers for image recognition at scale,'' 2020. [Online]. Available: \url{https://arxiv.org/abs/2010.11929}
\BIBentrySTDinterwordspacing

\bibitem{Lin2021}
\BIBentryALTinterwordspacing
T.~Lin, Y.~Wang, X.~Liu, and X.~Qiu, ``A survey of transformers,'' 2021. [Online]. Available: \url{https://arxiv.org/abs/2106.04554}
\BIBentrySTDinterwordspacing

\bibitem{Deng2018}
J.~Deng, J.~Guo, N.~Xue, and S.~Zafeiriou, ``Arcface: Additive angular margin loss for deep face recognition,'' in \emph{Proceedings of the IEEE/CVF Conference on Computer Vision and Pattern Recognition}, 2019, pp. 4690--4699.

\bibitem{Gray2007EvaluatingAM}
\BIBentryALTinterwordspacing
D.~Gray, S.~Brennan, and H.~Tao, ``Evaluating appearance models for recognition, reacquisition, and tracking,'' in \emph{Proc. IEEE international workshop on performance evaluation for tracking and surveillance (PETS)}, vol.~3, no.~5, 2007, pp. 1--7. [Online]. Available: \url{https://api.semanticscholar.org/CorpusID:15225312}
\BIBentrySTDinterwordspacing

\bibitem{Zhao2014}
\BIBentryALTinterwordspacing
R.~Zhao, W.~Ouyang, and X.~Wang, ``Learning mid-level filters for person re-identification,'' in \emph{2014 IEEE Conference on Computer Vision and Pattern Recognition}.\hskip 1em plus 0.5em minus 0.4em\relax IEEE, Jun. 2014. [Online]. Available: \url{http://dx.doi.org/10.1109/CVPR.2014.26}
\BIBentrySTDinterwordspacing

\bibitem{Schroff2015}
F.~Schroff, D.~Kalenichenko, and J.~Philbin, ``Facenet: A unified embedding for face recognition and clustering,'' in \emph{2015 IEEE Conference on Computer Vision and Pattern Recognition (CVPR)}, 2015, pp. 815--823.

\bibitem{liu2015faceattributes}
Z.~Liu, P.~Luo, X.~Wang, and X.~Tang, ``Deep learning face attributes in the wild,'' in \emph{Proceedings of International Conference on Computer Vision (ICCV)}, December 2015, pp. 3730--3738.

\end{thebibliography}

\end{document}